\documentclass[runningheads]{llncs}
\usepackage{graphicx}
\usepackage{amsmath,amssymb} 
\usepackage{color}
\usepackage{multirow}
\usepackage{subfig}
\usepackage[table]{xcolor}
\def\etal{\emph{et~al}.}
\def\eg{\emph{e.g.}}
\def\etc{\emph{etc}}
\def\ie{\emph{i.e.}}

\begin{document}
\pagestyle{headings}
\mainmatter

\def\ACCV18SubNumber{163}

\title{SAFE: Scale Aware Feature Encoder\\ for Scene Text Recognition} 
\titlerunning{SAFE: Scale Aware Feature Encoder for Scene Text Recognition}
\authorrunning{W. Liu et al.}

\author{Wei Liu \and Chaofeng Chen \and Kwan-Yee K. Wong}
\institute{Department of Computer Science, The University of Hong Kong\\
\email{\{wliu, cfchen, kykwong\}@cs.hku.hk}}

\maketitle

\begin{abstract}
In this paper, we address the problem of having characters with different scales in scene text recognition. We propose a novel scale aware feature encoder (SAFE) that is designed specifically for encoding characters with different scales. SAFE is composed of a multi-scale convolutional encoder and a scale attention network. The multi-scale convolutional encoder targets at extracting character features under multiple scales, and the scale attention network is responsible for selecting features from the most relevant scale(s). SAFE has two main advantages over the traditional single-CNN encoder used in current state-of-the-art text recognizers. First, it explicitly tackles the scale problem by extracting scale-invariant features from the characters. This allows the recognizer to put more effort in handling other challenges in scene text recognition, like those caused by view distortion and poor image quality. Second, it can transfer the learning of feature encoding across different character scales. This is particularly important when the training set has a very unbalanced distribution of character scales, as training with such a dataset will make the encoder biased towards extracting features from the predominant scale. To evaluate the effectiveness of SAFE, we design a simple text recognizer named scale-spatial attention network (S-SAN) that employs SAFE as its feature encoder, and carry out experiments on six public benchmarks. Experimental results demonstrate that S-SAN can achieve state-of-the-art (or, in some cases, extremely competitive) performance without any post-processing. 
\end{abstract}

\section{Introduction}

Scene text recognition refers to recognizing a sequence of characters that appear in a natural image. Inspired by the success \cite{bahdanau2014neural} in neural machine translation, many of the recently proposed scene text recognizers \cite{shi2016robust,lee2016recursive,cheng2017focusing,liu18aaai,cheng2017arbitrarily} adopt an encoder-decoder framework with an attention mechanism. Despite the remarkable results reported by them, very few of them have addressed the problem of having characters with different scales in the image. This problem often prevents existing text recognizers from achieving better performance. 

\begin{figure}[htb]
\begin{minipage}{\linewidth}
   \begin{center}
     \includegraphics[width=\linewidth]{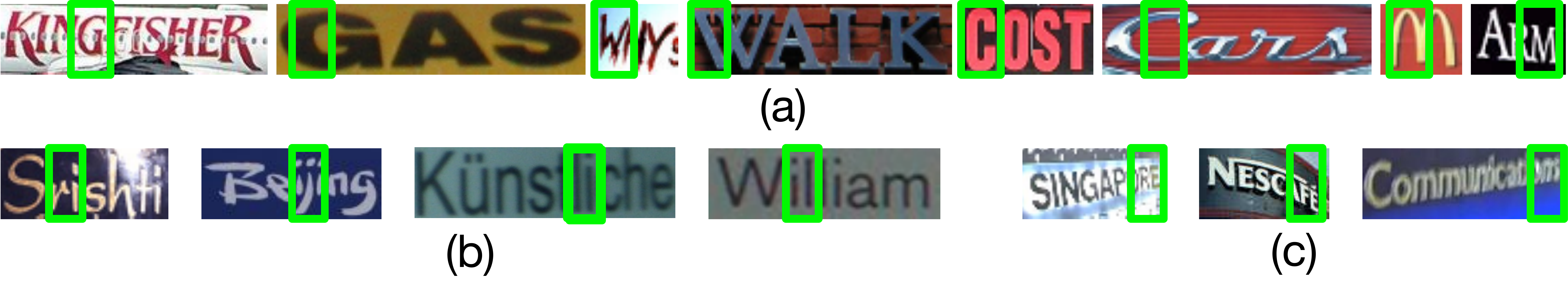}
   \end{center}
\end{minipage} 
\caption{Examples of text images having characters with different scales. The green rectangles represent the effective receptive field of a single-CNN encoder. (a) The character scale may vary greatly because of font style, font size, text arrangements, \etc. (b) For characters in the same image with the same font style and font size, their scales may still vary depending on which characters they are. For instance, lower case letters `l' and `i' are usually much `narrower' than other characters. (c) The distortion of text can also result in different character scales within the same image.}
\label{fig:examples}
\end{figure}

In a natural image, the scale of a character can vary greatly depending on which character it is (for instance, `M' and 'W' are in general wider than `i' and `l'), font style, font size, text arrangement, viewpoint, \etc. Fig.~\ref{fig:examples} shows some examples of text images containing characters with different scales. Existing text recognizers \cite{shi2016robust,lee2016recursive,cheng2017focusing,liu18aaai,cheng2017arbitrarily,shi2016end,liu16star} employ only one single convolutional neural network for feature encoding, and often perform poorly for such text images. Note that a single-CNN encoder with a fixed receptive field\footnote{Although the receptive field of a CNN is large, its effective region \cite{luo2016understanding} responsible for calculating each feature representation only occupies a small fraction.} (refer to the green rectangles in Fig.~\ref{fig:examples}) can only effectively encode characters which fall within a particular scale range. For characters outside this range, the encoder captures either only partial context information of a character or distracting information from the cluttered background and neighboring characters. In either case, the encoder cannot extract discriminative features from the corresponding characters effectively, and the recognizer will have great difficulties in recognizing such characters. 

\vspace{-0.3cm}
\vspace{-0.3cm}

\begin{figure}[htb]
\begin{minipage}{\linewidth}
	\begin{minipage}{\linewidth}
   \begin{center}
      \subfloat[]{\label{subfig:complicated_examples}\includegraphics[width=\linewidth]{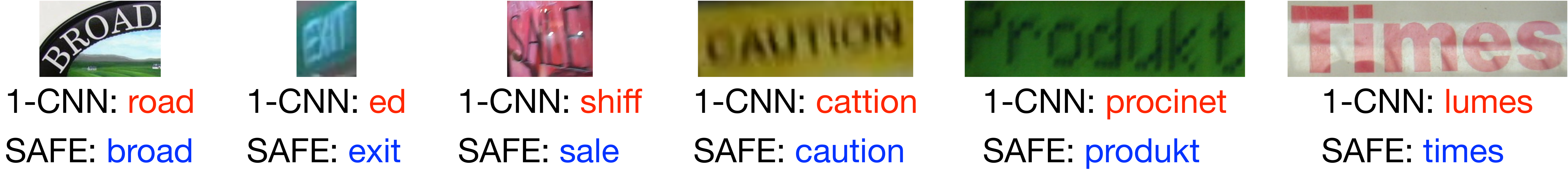}}
   \end{center}
   \end{minipage}
   \begin{minipage}{0.5\linewidth}
   \begin{center}
      \subfloat[]{\label{subfig:dataset_distribution}\includegraphics[width=\linewidth]{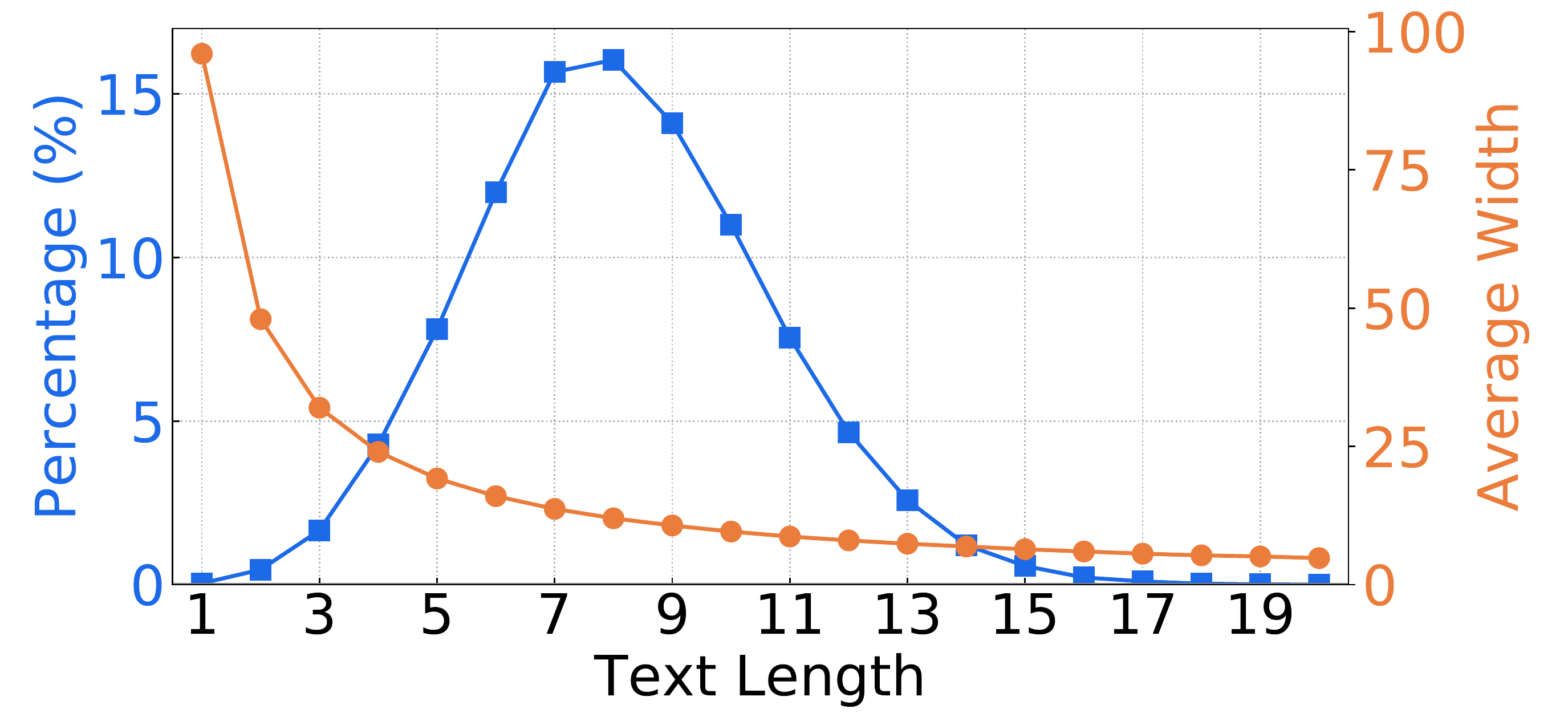}}
   \end{center}
   \end{minipage}
   \begin{minipage}{0.5\linewidth}
   \begin{center}
      \subfloat[]{\label{subfig:detailed_performance}\includegraphics[width=\linewidth]{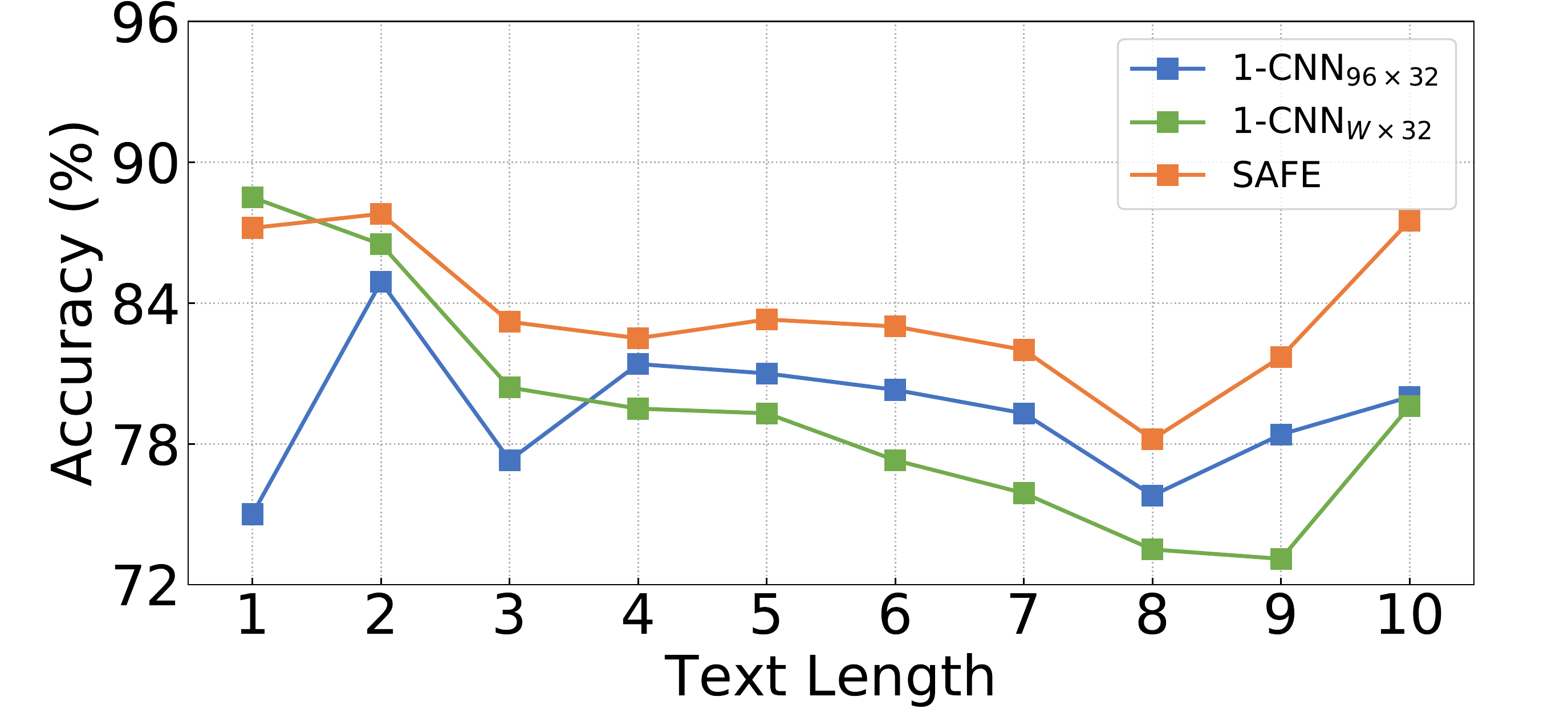}}
   \end{center}
   \end{minipage}
\end{minipage} 
\caption{Advantages of SAFE over the single-CNN encoder (1-CNN). (a) SAFE performs better than 1-CNN in handling distorted or low-quality text images. (b) The distribution of text lengths in SynthR \cite{jaderberg14syntheticdata} and the average character width in text images of different lengths after being resized to $96\times32$. (c) Detailed performance of recognizers with 1-CNN and with SAFE on recognizing text of different lengths. All the models are trained using SynthR. 1-CNN$_{96\times32}$ is trained by resizing the images to $96\times32$, whereas 1-CNN$_{W\times32}$ is trained by only rescaling images to a fixed height while keeping their aspect ratios unchanged.}
\label{fig:unbalanced_training_dataset}
\end{figure}

In this paper, we address the problem of having characters with different scales by proposing a simple but efficient scale aware feature encoder (SAFE). SAFE is designed specifically for encoding characters with different scales. It is composed of (i) a multi-scale convolutional encoder for extracting character features under multiple scales, and (ii) a scale attention network for automatically selecting features from the most relevant scale(s). SAFE has two main advantages over the traditional single-CNN encoder used in current state-of-the-art text recognizers. First, it explicitly tackles the scale problem by extracting scale-invariant features from the characters. This allows the recognizer to put more effort in handling other challenges in scene text recognition, like those caused by character distortion and poor image quality (see Fig.~\ref{subfig:complicated_examples}). Second, it can transfer the learning of feature encoding across different character scales. This is particularly important when the training set has a very unbalanced distribution of character scales. For instance, the most widely adopted SynthR dataset \cite{jaderberg14syntheticdata} has only a small number of images containing a single character (see Fig.~\ref{subfig:dataset_distribution}). In order to keep the training and testing procedures simple and efficient, most of the previous methods \cite{shi2016robust,lee2016recursive,cheng2017focusing,cheng2017arbitrarily,liu16star} resized an input image to a fixed resolution before feeding it to the text recognizer. This resizing operation will therefore result in very few training images containing large characters. Obviously, a text recognizer with a single-CNN encoder cannot learn to extract discriminative features from large characters due to limited training examples. This leads to a poor performance on recognizing text images with a single character (see Fig.~\ref{subfig:detailed_performance}). Alternatively, \cite{shi2016end,ijcai2017-458} resized the training images to a fixed height while keeping their aspect ratios unchanged. This, however, can only partially solve the scale problem as character scales may still vary because of font style, text arrangement, distortion, \etc. Hence, it is still unlikely to have a well-balanced distribution of character scales in the training set. Training with such a dataset will definitely make the encoder biased towards extracting features from the predominant scale and not able to generalize well to characters of different scales. Unlike the single-CNN encoder, SAFE can share the knowledge of feature encoding across different character scales and effectively extract discriminative features from characters of different scales. This enables the text recognizer to have a much more robust performance (see Fig.~\ref{subfig:detailed_performance}). 

\vspace{-0.3cm}
\vspace{-0.3cm}

To evaluate the effectiveness of SAFE, we design a simple text recognizer named scale-spatial attention network (S-SAN) that employs SAFE as its feature encoder. Following \cite{liu18aaai,ijcai2017-458}, S-SAN employs a spatial attention network in its LSTM-based decoder to handle a certain degree of text distortion. Experiments are carried out on six public benchmarks, and experimental results demonstrate that S-SAN can achieve state-of-the-art (or, in some cases, extremely competitive) performance without any post-processing. 

\section{Related Work}
Many methods have been proposed for scene text recognition in recent years. According to their strategies of recognizing words in text images, previous text recognizers can be roughly classified into bottom-up and top-down approaches. 

Recognizers based on the bottom-up approach first perform individual character detection and recognition using character-level classifiers, and then they integrate the results across the whole text image to get the final word recognition. In \cite{wang2010word,wang2011end,neumann2012real,yao2014strokelets}, traditional hand-crafted features (\eg, HOG) were employed to train classifiers for detecting and recognizing characters in the text image. With the recent success of deep learning, many methods \cite{wang2012end,bissacco2013photoocr,jaderberg14deepfeatures,Jaderberg15deepstructured} used deep neural networks to train character classifiers. In \cite{wang2012end}, Wang \etal~employed a CNN to extract features from the text image for character recognition. 
\cite{bissacco2013photoocr} first over-segmented the whole text image into multiple character regions before using a fully-connected neural network for character recognition. 
Unlike \cite{wang2012end,bissacco2013photoocr} which used a single character classifier, \cite{jaderberg14deepfeatures} combined a binary text/no-text classifier, a character classifier and a bi-gram classifier to compute scores for candidate words in a fixed lexicon. 
To recognise text without a lexicon, a structured output loss was used in their extended work \cite{Jaderberg15deepstructured} to optimize the individual character and $N$-gram predictors.

The above bottom-up methods require the segmentation of each character, which is a highly non-trivial task due to the complex background and different font size of text in the image. Recognizers \cite{shi2016robust,lee2016recursive,cheng2017focusing,cheng2017arbitrarily,shi2016end,liu16star,jaderberg14syntheticdata,ijcai2017-458,jaderberg2016reading,He2016ReadingST} which are based on the top-down approach directly recognize the entire word in the image. In \cite{jaderberg14syntheticdata} and \cite{jaderberg2016reading}, Jaderberg \etal~extracted CNN features from the entire image, and performed a 90k-way classification (90k being the size of a pre-defined dictionary). Instead of using only CNNs, \cite{shi2016end,liu16star,He2016ReadingST} also employed recurrent neural networks to encode features of word images. All these three models were optimized by the connectionist temporal classification \cite{graves2006connectionist}, which does not require an alignment between the sequential features and the ground truth labelling. 

Following the success of the attention mechanism \cite{bahdanau2014neural} in neural machine translation, recent text recognizers \cite{shi2016robust,lee2016recursive,cheng2017focusing,liu18aaai,cheng2017arbitrarily,ijcai2017-458} introduced a learnable attention network in their decoders to automatically select the most relevant features for recognizing individual characters. In order to handle distorted scene text, \cite{shi2016robust,liu16star} employed a spatial transformer network (STN) \cite{jaderberg15spatial} to rectify the distortion of the entire text image before recognition. As it is difficult to successfully train a STN from scratch, Cheng \etal~\cite{cheng2017arbitrarily} proposed to encode the text image from multiple directions and used a filter gate to generate the final features for decoding. Unlike \cite{shi2016robust,cheng2017arbitrarily,liu16star} which rectified the distortion of the entire image, a hierarchical attention mechanism was introduced in \cite{liu18aaai} to rectify the distortion of individual characters. 

Different from \cite{shi2016robust,liu18aaai,cheng2017arbitrarily,liu16star} which focused on recognizing severely distorted text images, we address the problem of having character with different scales. This is a common problem that exists in both distorted and undistorted text recognition. The scale-spatial attention network (S-SAN) proposed in this paper belongs to the family of attention-based encoder-decoder neural networks. Unlike previous methods \cite{shi2016robust,lee2016recursive,cheng2017focusing,liu18aaai,cheng2017arbitrarily,shi2016end,liu16star} which employed only a single CNN as their feature encoder, we introduce a scale aware feature encoder (SAFE) to extract scale-invariant features from characters with different scales. This guarantees a much more robust feature extraction for the text recognizer. Although a similar idea has been proposed for semantic segmentation \cite{chen2016attention} to merge segmentation maps from different scales, it is the first time that the scale problem has been identified and efficiently handled for text recognition using an attention-based encoder-decoder framework. Better still, SAFE can also be easily deployed in other text recognizers to further boost their performance.

\section{Scale Aware Feature Encoder}\label{section:safe}
The scale aware feature encoder (SAFE) is conceptually simple: in order to extract discriminative features from characters with different scales, feature encoding of the text image should first be carried out under different scales, and features from the most relevant scale(s) are then selected at each spatial location to form the final feature map for the later decoding. SAFE is thus a natural and intuitive idea. As illustrated in Fig.~\ref{figure:safe}, SAFE is composed of a multi-scale convolutional encoder and a scale-attention network. Throughout this paper, we omit the bias terms for improved readability.

\vspace{-0.2cm}
\subsubsection{Multi-Scale Convolutional Encoder.}
\begin{figure*}[t]
\begin{minipage}{\linewidth}
      \begin{center}
   \begin{minipage}{\linewidth}
      \begin{center}
      \includegraphics[width=\linewidth]{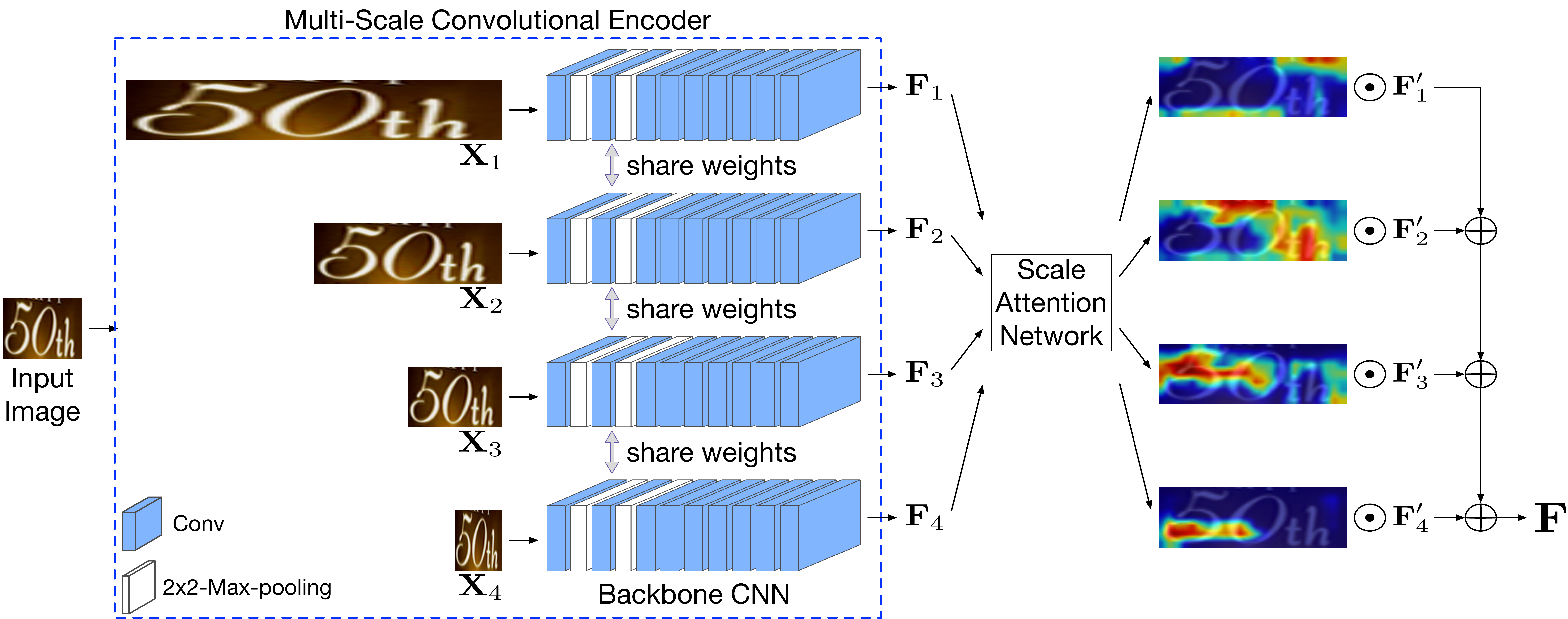}
      \end{center}
   \end{minipage}
          \end{center}
   \end{minipage}
   \caption{The architecture of SAFE ($N=4$). As most of the text images contain only a single word, the scale problem along the height dimension is not as severe as that along the width dimension. In our implementation, we resize the input images to 4 different resolutions, namely $192\times32$, $96 \times 32$, $48 \times 32$ and $24 \times 32$ respectively. The saliency maps in the scale attention network represent the scale attention for the feature maps extracted under 4 different resolutions. $\odot$ and $\oplus$ denote element-wise multiplication and summation respectively. }
\label{figure:safe}
\end{figure*}

The multi-scale convolutional encoder is responsible for encoding the original text images under multiple scales. The basic component of the multi-scale convolutional encoder takes the form of a backbone convolutional neural network. Given a single gray image $\mathbf{I}$ as input, a multi-scale pyramid of images $\{\mathbf{X}_s\}_{s=1}^{N}$ is first generated. $\mathbf{X}_s$ here denotes the image at scale $s$ having a dimension of ${W_s \times H_s}$ (width$\times$height), and $\mathbf{X}_1$ is the image at the finest scale. In order to encode each image in the pyramid, $N$ backbone CNNs which share the same set of parameters are employed in the multi-scale convolutional encoder (see Fig.~\ref{figure:safe}). The output of our multi-scale convolutional encoder is a set of $N$ spatial feature maps
\begin{equation}
\{\mathbf{F}_1, \mathbf{F}_2, \cdots, \mathbf{F}_N\} = \{{\rm CNN}_1(\mathbf{X}_1), {\rm CNN}_2(\mathbf{X}_2), \cdots, {\rm CNN}_N(\mathbf{X}_N)\},
\end{equation}
where ${\rm CNN}_s$ and $\mathbf{F}_s$ denote the backbone CNN and the output feature map at scale $s \in \{1, 2, \cdots, N\}$ respectively. $\mathbf{F}_s$ has a dimension of $W'_s \times H'_s \times C'$ (width$\times$height$\times$\#channels). 

Unlike dominant backbone CNNs adopted by \cite{shi2016robust,cheng2017focusing,cheng2017arbitrarily,shi2016end,liu16star} which compress information along the image height dimension and generate unit-height feature maps, each of our backbone CNNs keeps the spatial resolution of its feature map $\mathbf{F}_s$ close to that of the corresponding input image $\mathbf{X}_s$. Features from both text and non-text regions, as well as from different characters, therefore remain distinguishable in both the width and height dimensions in each feature map. In order to preserve the spatial resolution, we only apply two $2\times2$ down-sampling layers (see Fig.~\ref{figure:safe}) in each of our backbone CNNs. Each feature map $\mathbf{F}_s$ is therefore sub-sampled only four times in both width and height (i.e., $W_s' = W_s/4$ and $H_s' = H_s/4$). More implementation details related to our backbone CNNs can be found in Section~\ref{subsection:network_architecture}.

\vspace{-0.2cm}
\subsubsection{Scale Attention Network.} The scale attention network is responsible for automatically selecting features from the most relevant scale(s) at each spatial location to generate the final scale-invariant feature map $\mathbf{F}$. In the proposed scale attention network, the $N$ convolutional feature maps $\{\mathbf{F}_1, \mathbf{F}_2, \cdots, \mathbf{F}_N\}$ outputted from the multi-scale convolutional encoder are first up-sampled to the same resolution
\begin{equation}
\{\mathbf{F}'_1, \mathbf{F}'_2, \cdots, \mathbf{F}'_N\} = \{u(\mathbf{F}_1), u(\mathbf{F}_2), \cdots, u(\mathbf{F}_N)\},
\end{equation}
where $u(\cdot)$ is an up-sampling function, and $\mathbf{F}'_s$ is the up-sampled feature map with a dimension of $W' \times H' \times C'$ (width$\times$height$\times$\#channels). Through extensive experiments, we find that the performances of different up-sampling functions (\eg, bilinear interpolation, nearest interpolation, \etc.) are quite similar. We therefore simply use bilinear interpolation in our implementation.

\begin{figure*}[t]
\begin{minipage}{\linewidth}
      \begin{center}
   \begin{minipage}{\linewidth}
      \begin{center}
      \includegraphics[width=0.95\linewidth]{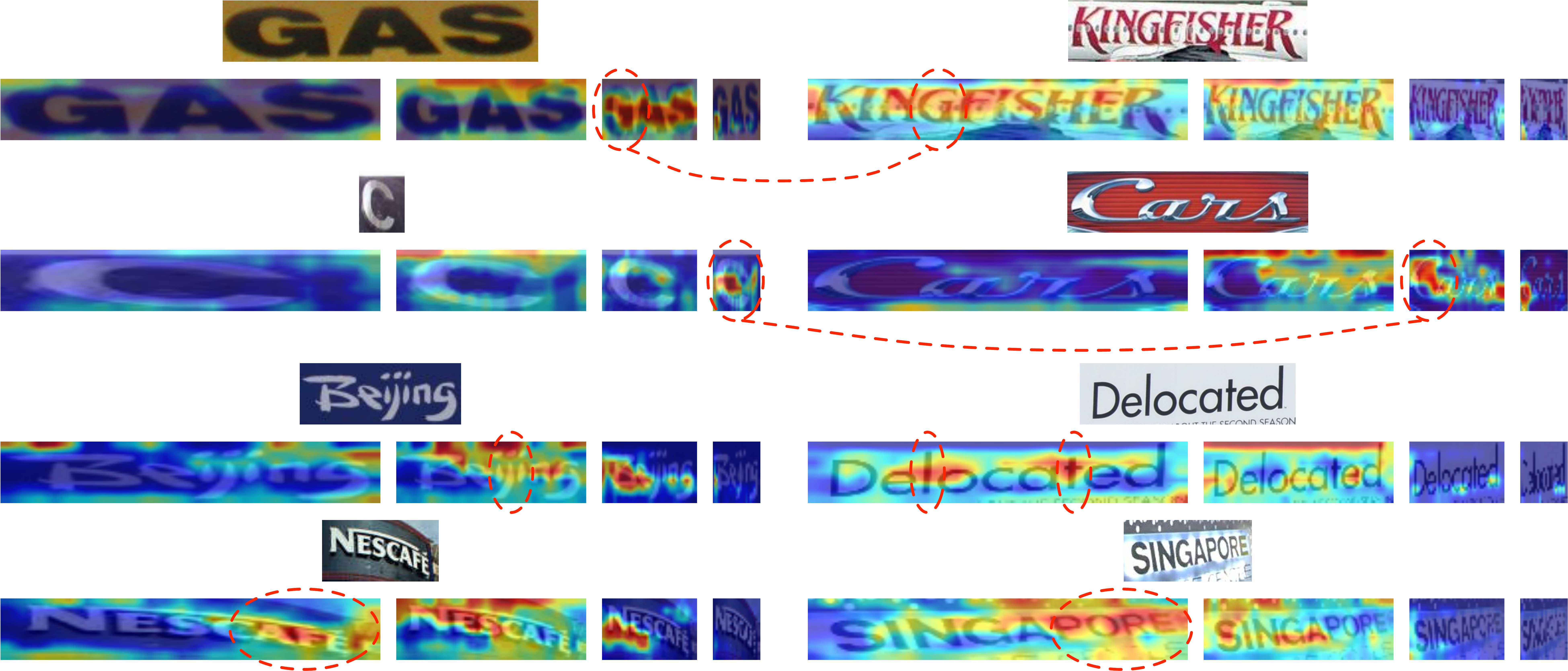}
      \end{center}
   \end{minipage}
          \end{center}
   \end{minipage}
   \caption{Visualization of scale attention in the proposed SAFE. The saliency maps of the scale attention are superimposed on the images of the corresponding scales. }
\label{figure:examples_safe}
\end{figure*}

After obtaining the up-sampled feature maps, our scale attention network utilizes an attention mechanism to learn to weight features from different scales. Since the size of the characters may not be constant within a text image (see Fig.~\ref{figure:safe}), the computation of attention towards features from different scales takes place at each spatial location. In order to select the features from the most relevant scale(s) at the spatial location $(i, j)$, $N$ scores are first computed for evaluating the importance of features from the $N$ different scales
\begin{equation}
	\begin{bmatrix}
		f_1(i,j) \\
		f_2(i,j) \\
		\vdots \\
		f_N(i,j) \\
   	\end{bmatrix} = \mathbf{W}\begin{bmatrix}\mathbf{F}'_1(i,j)\\\mathbf{F}'_2(i,j)\\\vdots\\\mathbf{F}'_N(i,j)\\\end{bmatrix}\label{equation:scale_attention_score}
\end{equation}
where $f_s(i,j)$ is the score of the $C'$-dimensional feature vector $\mathbf{F}'_s(i,j)$ at location $(i, j)$ of the corresponding feature map $\mathbf{F}'_s$ at scale $s$, and $\mathbf{W}$ is the parameter matrix. The proposed scale attention network then defines the scale attention at location $(i,j)$ as
\begin{equation}\label{equation:scale_attention}
\omega_s(i,j) = \frac{\exp(f_s(i,j))}{\sum_{s'=1}^{N}\exp(f_{s'}(i,j))},
\end{equation}
where $\omega_s(i,j)$ is the attention weight for the feature vector $\mathbf{F}'_s(i,j)$. Finally, the scale-invariant feature map $\mathbf{F}$ can be computed as
\begin{equation}
  \mathbf{F}(i,j) = \sum^N_{s=1}\omega_s(i,j)\mathbf{F}'_s(i,j).
\end{equation}
The final feature map $\mathbf{F}$ has the same dimension as each up-sampled feature map $\mathbf{F}'_s$. Note that the scale attention network together with the multi-scale convolutional encoder are optimized in an end-to-end manner using only the recognition loss. As illustrated in Fig.~\ref{figure:examples_safe}, although we do not have the ground truth to supervise the feature selection across different scales, the scale attention network can automatically learn to attend on the features from the most relevant scale(s) for generating the final feature map $\mathbf{F}$.

\section{Scale-Spatial Attention Network}

\begin{figure*}[t]
   \begin{minipage}{\linewidth}
      \centering
      \includegraphics[width=\linewidth]{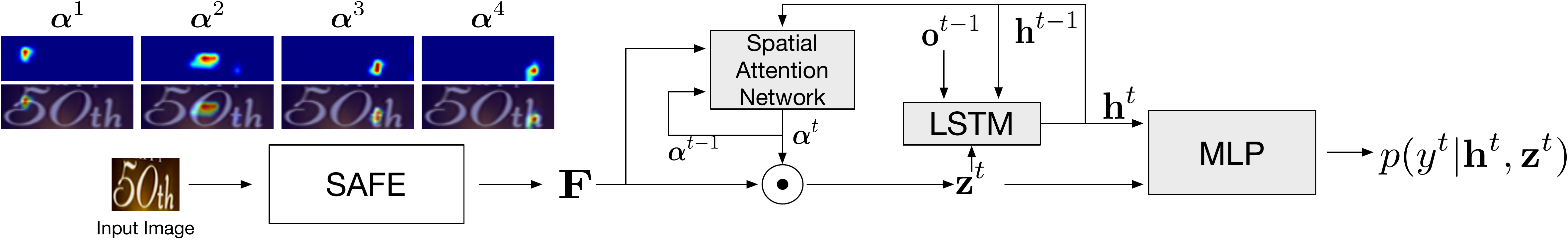}
   \end{minipage}
   \caption{The architecture of S-SAN. The saliency maps represent the spatial attention at different decoding steps. $\odot$ denotes element-wise multiplication. }
\label{figure:overall_architect}
\end{figure*}

To demonstrate the effectiveness of SAFE proposed in Section~\ref{section:safe}, we design a scale-spatial attention network (S-SAN) for scene text recognition. S-SAN uses SAFE as its feature encoder, and employs a character aware decoder composed of a spatial attention network and a LSTM-based decoder. Fig.~\ref{figure:overall_architect} shows the overall architecture of S-SAN. 

\subsection{SAFE}\label{subsection:sace}

To extract scale-invariant features for recognizing characters with different scales, S-SAN first employs SAFE proposed in Section~\ref{section:safe} to encode the original text image. By imposing weight sharing across backbone CNNs of different scales, SAFE keeps its number of parameters to a minimum. Comparing with previous encoders adopted by \cite{shi2016robust,lee2016recursive,cheng2017focusing,liu18aaai,cheng2017arbitrarily,shi2016end,liu16star,bai2018edit} for text recognition, SAFE not only keeps its structure as simple as possible (see Table~\ref{table:architecture}), but also effectively handles characters with different scales. This enables S-SAN to achieve a much better performance on recognizing text in natural images (see Table~\ref{table:accuracies_on_six_benchmarks} and Table~\ref{table:accuracies_synthm}).

\subsection{Character Aware Decoder}\label{subsection:cad}

The character aware decoder of S-SAN is responsible for recurrently translating the scale-invariant feature map $\mathbf{F}$ into the corresponding ground truth labelling $\mathbf{y} = \{y^1,y^2,...,y^{T}, y^{T+1}\}$. Here, $T$ denotes the length of the text and $y^{T+1}$ is the end-of-string (eos) token representing the end of the labelling. In this section, we refer to the process of predicting each character $y^t$ as one time/decoding step. 

\subsubsection{Spatial Attention Network.}\label{section:span}

The function of the spatial attention network is to generate a sequence of context vectors from the scale-invariant feature map $\mathbf{F}$ for recognizing individual characters. Following \cite{liu18aaai,ijcai2017-458}, our spatial attention network extends the standard 1D attention mechanism \cite{shi2016robust,lee2016recursive,cheng2017focusing,cheng2017arbitrarily} to the 2D spatial domain. This allows the decoder to focus on features at the most relevant spatial locations for recognizing individual characters, and handle a certain degree of text distortion in natural images. At a particular decoding step $t$, we compute the context vector $\mathbf{z}^t$ as a weighted sum of all feature vectors in $\mathbf{F}$, i.e., 
\begin{equation}\label{equation:attention_weighted_sum}
\mathbf{z}^t = \sum_{i=1}^{W'}\sum_{j=1}^{H'}\alpha^t(i,j)\mathbf{F}(i,j),
\end{equation}
where $\alpha^t(i,j)$ is the weight applied to $\mathbf{F}(i,j)$ as determined by the spatial attention network. For the recognition of a single character at decoding step $t$, the context vector $\mathbf{z}^t$ should only focus on features at the most relevant spatial locations (i.e., those  corresponding to the single character being recognized). To compute $\alpha^t(i,j)$ at the current decoding step $t$ , we first exploit a simple CNN to encode the previous attention map $\boldsymbol{\alpha}^{t-1}$ into
\begin{equation} \label{equation:attention_cnn_encode_previous_attention_maps}
  \mathbf{A}^{t-1} = {\rm CNN}_{\rm spatial}(\boldsymbol{\alpha}^{t-1}).
\end{equation}
We then evaluate a relevancy score at each spatial location as
\begin{equation}\label{equation:attention_score}
  r^t(i,j) = \mathbf{w}^{\rm T}{\rm tanh}[\mathbf{M}\mathbf{h}^{t-1}+ \mathbf{U}\mathbf{A}^{t-1}(i,j)+ \mathbf{V}\mathbf{F}(i,j)],
\end{equation}
where $\mathbf{h}^{t-1}$ is the previous hidden state of the decoder (explained later), $\mathbf{M}$, $\mathbf{U}$ and $\mathbf{V}$ are the parameter matrices, and $\mathbf{w}$ is a parameter vector. Finally, we normalize the scores to obtain the attention weight
\begin{equation}\label{equation:attention_softmax}
  \alpha^t(i,j) = \frac{\exp(r^t(i,j))}{\sum_{i'=1}^{W'}\sum_{j'=1}^{H'}\exp(r^t(i',j'))}.
\end{equation}


\subsubsection{LSTM-based Decoder.}\label{section:lstm_based_decoder}
The actual decoder of S-SAN takes the form of a long short-term memory layer and a multi-layer perceptron (MLP). Let $L$ denote the set of 37-class (26 letters + 10 digits + eos) case-insensitive characters in our task. At the decoding step $t$, the LSTM-based decoder defines a probability distribution over $L$ as
\begin{align}
\mathbf{h}^{t} &= {\rm LSTM}(\mathbf{h}^{t-1}, \mathbf{o}^{t-1}, \mathbf{z}^t)\label{equation:decoder_lstm}\\
p(y^t|\mathbf{h}^{t}, \mathbf{z}^t) &= {\rm SoftMax}(\mathbf{W}_y\begin{bmatrix}\mathbf{h}^{t}\\{\rm tanh}(\mathbf{W}_z\mathbf{z}^{t})\\\end{bmatrix})\label{equation:decoder_mlp},
\end{align} 
where $\mathbf{h}^{t-1}$ and $\mathbf{h}^{t}$ denote the previous and current hidden states respectively, $\mathbf{o}^{t-1}$ is the one-hot encoding of the previous character $y^{t-1}$, $\mathbf{W}_y$ and $\mathbf{W}_z$ are the parameters of two linear layers in the MLP, and ${\rm SoftMax}$ denotes the final layer of the MLP for outputting the probability. 
The probability of the sequential labelling is then given by the product of the probability of each label, i.e., 
\begin{equation}\label{equation:decoder_probability_word}
p(\mathbf{y}|\mathbf{I}) = \prod_{t=1}^{T+1}p(y^t|\mathbf{h}^{t}, \mathbf{z}^t).
\end{equation}

\section{Datasets and Implementation Details}


\subsection{Datasets}\label{subsection:datasets}

Following \cite{shi2016robust,lee2016recursive,liu18aaai,shi2016end,liu16star,Jaderberg15deepstructured,jaderberg2016reading}, S-SAN is trained using the most commonly adopted dataset released by Jaderberg \etal~\cite{jaderberg14syntheticdata} (referred to as SynthR). This dataset contains 8-million synthetic text images together with their corresponding word labels. The distribution of text lengths in SynthR is shown in Fig.~\ref{subfig:dataset_distribution}. The evaluation of S-SAN is carried out on the following six public benchmarks. 
\begin{itemize}
	\item \textbf{IIIT5K}~\cite{mishra2012scene} contains 3,000 word test images collected from the Internet. Each image has been associated to a 50-word lexicon and a 1K-word lexicon.
	\item \textbf{Street View Text (SVT)}~\cite{wang2011end} contains 647 test word images which are collected from Google Street View. Each word image has a 50-word lexicon.
	\item \textbf{ICDAR-2003 (IC03)}~\cite{lucas2005icdar} contains $860$ text images for testing. Each image has a 50-word lexicon defined by Wang \etal~\cite{wang2011end}. A full lexicon is constructed by merging all 50-word lexicons. Following \cite{wang2011end}, we recognize the images having only alphanumeric words (0-9 and A-Z) with at least three characters. 
	\item \textbf{ICDAR-2013 (IC13)}~\cite{karatzas2013icdar} is derived from ICDAR-2003, and contains 1,015 cropped word test images without any pre-defined lexicon. Previous methods \cite{shi2016robust,lee2016recursive,liu18aaai} recognise images containing only alphanumeric words with at least three characters, which results in 857 images left for evaluation. In Section \ref{section:experiment}, we refer to IC13 with 1,015 images as {\bf IC13-L} and that with 857 images as {\bf IC13-S}, where ‘L’ and ‘S’ stand for large and small respectively.
	\item \textbf{Street View Text Perspective (SVT-P)}~\cite{phan2013recognizing} contains 639 test images which are specially picked from the side-view angles in Google Street View. Most of them suffer from a large perspective distortion. 
	\item \textbf{ICDAR Incidental Scene Text (ICIST)}~\cite{karatzas2015icdar} contains 2,077 text images for testing. Many of them are severely distorted and blurry. To make a fair comparison with \cite{cheng2017focusing,bai2018edit}, we also discard the images with non-alphanumeric characters in ICIST, which results in 1,811 images for testing. Like IC13, we refer to ICIST with 2,077 images as {\bf ICIST-L} and that with 1,811 images as {\bf ICIST-S} in Section \ref{section:experiment}.
\end{itemize}

\subsection{Network Architecture}\label{subsection:network_architecture}

\subsubsection{Scare Aware Feature Encoder.} The detailed architecture of the backbone CNN in SAFE is illustrated in Table \ref{table:architecture}.  The basic backbone CNN consists of nine convolutional layers. The stride and padding size of each convolutional layer are both equal to 1. All the convolutional layers use ReLU as the activation function. Batch normalization \cite{ioffe2015batch} is used after every convolutional layer. 
In our implementation, the multi-scale pyramid of images used in the multi-scale convolutional encoder has 4 scales, with images having a dimension of $192\times32$, $96 \times 32$, $48 \times 32$ and $24 \times 32$ respectively. The dimensions of the corresponding feature maps $\{\mathbf{F}_s\}_{s=1}^{4}$ are $48\times8\times512$, $24\times8\times512$, $12\times8\times512$ and $6\times8\times512$ respectively. In our scale attention network, the spatial resolutions of the up-sampled\footnote{We obtain $\textbf{F}_1'$ by actually down-sampling $\textbf{F}_1$.} feature maps $\{\mathbf{F}_s'\}_{s=1}^{4}$ are all $24\times8$ (i.e., same as that of $\mathbf{F}_2$ which is extracted from $\textbf{X}_2$ with a dimension of $96\times32$). 

\begin{table*}[!t]
\begin{center}
\tabcolsep=0.02cm
\renewcommand{\arraystretch}{1.1}
\setlength{\tabcolsep}{.13em}
\caption{Comparison of encoders among different text recognizers. `STN', `BLSTM' and `SCAN' denote the spatial transformer network \cite{jaderberg15spatial}, the bi-directional LSTM and our scale attention network respectively.}\label{table:architecture}
\scriptsize
\begin{tabular}{|l|c|c|c|c|c|c|c|c|c|}
\hline
\multirow{2}*{Model}         & \multirow{2}*{STN} & \multicolumn{5}{c|}{Backbone CNN}                                                 & \multirow{2}*{BLSTM} &\multirow{2}*{SCAN}& Encoder\\ 
\cline{3-7}
                             &                    & Unit 1        & Unit 2         & Unit 3         & Unit 4         & Unit 5         &                      &                   & Size\\
\hline
CRNN~\cite{shi2016end}                   & 0 & [$3, 64$]$\times1$ & [$3, 128$]$\times1$ & [$3, 256$]$\times2$ & [$3, 512$]$\times2$ & \multirow{2}*{[$2, 512$]$\times1$} & \multirow{2}*{2} & \multirow{2}*{0} & 8.3M\\
\cline{1-2}\cline{10-10}
RARE~\cite{shi2016robust}                & 1 & (2,2,2,2)          & (2,2,2,2)           & (2,2,1,2)           & (2,2,1,2)           &                                    &                  &                  & 16.5M \\
\hline
\multirow{2}*{STAR-Net~\cite{liu16star}} & \multirow{2}*{1} & \cellcolor{gray!30}[$3, 64$]$\times5$ & \cellcolor{gray!30}[$3, 128$]$\times4$ & \cellcolor{gray!30}[$3, 256$]$\times4$ & \cellcolor{gray!30}[$3, 512$]$\times4$ & \multirow{2}*{[$3, 512$]$\times1$} & \multirow{2}*{1} & \multirow{2}*{0} & \multirow{2}*{14.6M}\\
                                         &                  & (2,2,2,2)          & (2,2,2,2)           & (1,2,1,2)           & (1,2,1,2)           & &&&\\

\hline
\multirow{2}*{Char-Net~\cite{liu18aaai}} & \multirow{2}*{1} & [$3, 64$]$\times3$ & [$3, 128$]$\times2$ & \multirow{2}*{[$3, 256$]$\times6$} & \multirow{2}*{[$3, 256$]$\times4$} & \multirow{2}*{[$3, 512$]$\times3$} & \multirow{2}*{0} & \multirow{2}*{0} & \multirow{2}*{12.8M} \\
                                         &                  & (2,2,2,2)          & (2,2,2,2)           & & & & & &\\
\hline
\multirow{2}*{FAN~\cite{cheng2017focusing}} & \multirow{3}*{0} & [$3, 32$]$\times1$ & \cellcolor{gray!30} & \cellcolor{gray!30} & \cellcolor{gray!30} & \cellcolor{gray!30}& \multirow{3}*{1} & \multirow{3}*{0} & \multirow{3}*{45.8M}\\
                                            &                  & [$3, 64$]$\times1$ & \cellcolor{gray!30}\multirow{-2}*{[$3, 128$]$\times3$} & \cellcolor{gray!30}\multirow{-2}*{[$3, 256$]$\times5$} & \cellcolor{gray!30} & \cellcolor{gray!30}\multirow{-2}*{[$3, 512$]$\times6$} &&&\\
\multirow{-2}*{Bai \etal~\cite{bai2018edit}}                &                  & (2,2,2,2)          & (2,2,2,2)                          & (2,2,1,2)                          & \cellcolor{gray!30}\multirow{-3}*{[$3, 512$]$\times11$}                                    & [$2, 512$]$\times2$ &&& \\
\hline
1-CNN                & \multirow{2}*{0} & [$3, 64$]$\times1$ & [$3, 128$]$\times1$ & \multirow{2}*{[$3, 256$]$\times3$} & \multirow{2}*{[$3, 512$]$\times3$} & \multirow{2}*{[$3, 512$]$\times1$} & \multirow{2}*{0} & 0 & 9.03M \\
\cline{1-1}\cline{9-10}
SAFE &  & (2,2,2,2) & (2,2,2,2) &  &  &  & & 1 & 9.04M \\
\hline
\multirow{2}*{$\text{SAFE}_\text{res}$} & \multirow{2}*{0}                  & [$3, 64$]$\times2$ & \cellcolor{gray!30}[$3, 128$]$\times3$ & \cellcolor{gray!30} & \cellcolor{gray!30} & \cellcolor{gray!30} & \multirow{2}*{0}                    & \multirow{2}*{1}                 & \multirow{2}*{38.9M}\\
                                       &                                    & (2,2,2,2)          & (2,2,2,2) & \cellcolor{gray!30}\multirow{-2}*{[$3, 256$]$\times5$} &\cellcolor{gray!30}\multirow{-2}*{[$3, 512$]$\times9$}& \cellcolor{gray!30}\multirow{-2}*{[$3, 512$]$\times7$}&&&\\
\hline
\end{tabular}
\end{center}
\scriptsize
The configurations of all the convolutional layers in the backbone CNN follow the format of $[kernel~size, number~of~channels] \times number~of~layers$. Cells with a gray background indicate convolutional blocks with residue connections. For the max-pooling layers, the kernel size and strides in both width and height dimensions follow the format of $(kernel_{w}, kernel_{h}, stride_w, stride_h)$.
\end{table*}
\vspace{-0.3cm}
\subsubsection{Character Aware Decoder.} The character aware decoder employs a LSTM layer with 256 hidden states. In the spatial attention network, ${\rm CNN}_{\rm spatial}(\cdot)$ is one $7\times7$ convolutional layer with $32$ channels. In Eq.~(\ref{equation:attention_score}), $\mathbf{h}^{t-1}$, $\mathbf{A}(i,j)$ and $\mathbf{F}(i,j)$ are vectors with a dimension of $256$, $32$ and $512$ respectively, and $\mathbf{w}$ is a parameter vector of $256$ dimensions. Consequently, the dimensions of $\mathbf{M}$, $\mathbf{U}$ and $\mathbf{V}$ are $256\times256$, $256\times32$ and $256\times512$ respectively. The initial spatial attention map is set to zero at each position.

\subsection{Model Training and Testing.}\label{subsection:model_training_testing}

Adadelta \cite{zeiler2012adadelta} is used to optimize the parameters of S-SAN. During training, we use a batch size of 64. The proposed model is implemented using Torch7 and trained on a single NVIDIA GTX1080 GPU. It can process about 150 samples per second and converge in five days after about eight epochs over the training dataset. S-SAN is trained in an end-to-end manner only under the supervision of the recognition loss (i.e., the negative log-likelihood of Eq.(\ref{equation:decoder_probability_word})). During testing, for unconstrained text recognition, we directly pick the label with the highest probability in Eq.~(\ref{equation:decoder_mlp}) as the output of each decoding step. For constrained recognition with lexicons of different sizes, we calculate the edit distances between the prediction of the unconstrained text recognition and all words in the lexicon, and take the one with the smallest edit distance and highest recognition probability as our final prediction. 
In Table \ref{table:accuracies_on_six_benchmarks} and Table \ref{table:accuracies_synthm}, `$\rm None$' indicates unconstrained scene text recognition (\ie, without imposing any lexicon). `$50$', `$\rm 1K$' and `$\rm Full$' denote each scene text recognized with a 50-word lexicon, a 1,000-word lexicon and a full lexicon respectively.

\section{Experiments}\label{section:experiment}

Experiments are carried out on the six public benchmarks. Ablation study is first conducted in Section~\ref{subsection:ablation_study} to illustrate the effectiveness of SAFE. Comparison on the performance of S-SAN and other methods is then reported in Section \ref{subsection:comparative_evaluation}, which demonstrates that S-SAN can achieve state-of-the-art performance on standard benchmarks for text recognition. 

\subsection{Ablation Study}\label{subsection:ablation_study}
In this section, we conduct an ablation study to demonstrate the effectiveness of SAFE. Note that all the models in the ablation study are trained using the commonly adopted SynthR dataset, and we only use the accuracies of unconstrained text recognition for comparison.

\begin{table*}[t]
\renewcommand{\arraystretch}{1.1}
\centering
\caption{Ablation study of the proposed SAFE. }\label{table:ablation_study_safe}
\scriptsize
\begin{tabular}{|c|c|c|c|c|c|c|c|c|c|c|c|c|c|}
    \hline
    Model                                 & Image Size              & IIIT5K & SVT & IC03 & IC13-S  & IC13-L  & SVT-P & ICIST-S & ICIST-L \\
    \hline
    1-CNN$_{24\times32}$             & $24\times32$                 & 71.0   & 70.9& 82.8 & 80.2   & 79.3  &  54.6    & 56.1  & 51.0  \\
    \hline
    1-CNN$_{48\times32}$             & $48\times32$                 & 81.3   & 81.8& 90.5 & 87.6  & 86.2  &  68.8    & 66.3  & 60.3  \\
    \hline
    1-CNN$_{96\times32}$             & $96\times32$                 & 82.6   & 83.2& 91.0 & 89.7  & 87.7  &  69.9    & 67.1  & 61.7  \\
    \hline
    1-CNN$_{192\times32}$            & $192\times32$                & 82.1   & 83.2& 90.6 & 89.5  & 87.2  &  68.5    & 65.1  & 59.5  \\
    \hline
    1-CNN$_{W\times32}$              & $W\times32$                  & 83.6   & 82.2& 91.3 & 90.1  & 89.4  &  63.3    & 62.4  & 57.4  \\
    \hline
    \hline
   	\multirow{8}*{S-SAN}             & $48\times32$, $96\times32$   & 84.2   & 84.7& 92.3 & 89.7  & 88.3  &  71.6    & 70.5  & 64.4  \\
    \cline{2-10}
                                     & $96\times32$, $192\times32$  & 83.7   & 84.9& 92.4 & {\bf 91.1}  & 88.9  &  72.1    & 68.8  & 62.9  \\
    \cline{2-10}
                                     & $24\times32$, $48\times32$   &\multirow{2}*{85.0}&\multirow{2}*{84.4}&\multirow{2}*{92.0}&\multirow{2}*{90.7}&\multirow{2}*{89.8}&\multirow{2}*{72.9}&\multirow{2}*{70.7}&\multirow{2}*{64.7}\\
                                     & $96\times32$                 &        &     &      &       &       &       &&\\
   \cline{2-10}
                                     & $48\times32$, $96\times32$   &\multirow{2}*{85.0}&\multirow{2}*{84.5}&\multirow{2}*{91.9}&\multirow{2}*{90.7}&\multirow{2}*{89.7}&\multirow{2}*{71.8}&\multirow{2}*{70.3}&\multirow{2}*{64.4}\\
                                     & $192\times32$                &        &     &      &       &       &       &&\\
    \cline{2-10}
                                     & $24\times32$, $48\times32$   &\multirow{2}*{\bf 85.2}&\multirow{2}*{\bf 85.5}&\multirow{2}*{\bf 92.9}&\multirow{2}*{\bf 91.1}&\multirow{2}*{\bf 90.3}&\multirow{2}*{\bf 74.4}&\multirow{2}*{\bf 71.8}&\multirow{2}*{\bf 65.7}\\
                                     & $96\times32$, $192\times32$  &        &     &      &       &       &       &&\\
    \hline
\end{tabular}
\end{table*}

To demonstrate the advantages of SAFE over the single-CNN encoder (1-CNN), we first train five versions of text recognizers which employ single backbone CNN for feature encoding. The architecture of the backbone CNNs and the decoders in the single-scale recognizers are identical to those in S-SAN. As illustrated in Table \ref{table:ablation_study_safe}, we train the recognizers with the single-CNN encoder under different experimental settings, which resize the input text image to different resolutions. For simplicity, we directly use the encoder with the resolution of the training images to denote the corresponding recognizer. For example, 1-CNN$_{24\times32}$ stands for the single-CNN recognizer trained with images having a resolution of $24\times32$. As resizing images to a fixed resolution during training and testing would result in the training dataset having very few images containing large characters (see Fig.~\ref{subfig:dataset_distribution}), we also train a version of the single-CNN recognizer by resizing the image to a fixed height while keeping its aspect ratio unchanged (denoted as 1-CNN$_{W\times32}$ in Table~\ref{table:ablation_study_safe}). In order to train the recognizer with a batch size larger than 1, we normalize all the images to have a resolution of $192\times32$. For an image whose width is smaller than 192 after being resized to a height of 32, we pad it with the mean value to make its width equal to 192. Otherwise, we directly resize the image to $192\times32$ (there are very few text images with a width larger than 192). From the recognition results reported in Table \ref{table:ablation_study_safe}, we see that S-SAN with SAFE outperforms the other recognizers with a single-CNN encoder by a large margin. 

We also evaluate the scale selection of SAFE by using different combinations of rescaled images in the pyramid. As shown in Table \ref{table:ablation_study_safe}, S-SAN with four scales ($192\times32$, $96\times32$, $48\times32$ and $24\times32$) in the pyramid has the best performance on all the six public benchmarks.

\begin{table*}[t]
\tabcolsep=0.04cm
\renewcommand{\arraystretch}{1.1}
\begin{center}
\caption{Text recognition accuracies (\%) on six pubic benchmarks. *\cite{jaderberg2016reading} is not strictly unconstrained text recognition as its outputs are all constrained to a pre-defined 90K dictionary.}\label{table:accuracies_on_six_benchmarks}
\scriptsize
\begin{tabular}{|l|c|c|c|c|c|c|c|c|c|c|c|c|c|}
    \hline
    \multirow{2}*{Method}                            & \multicolumn{3}{c|}{IIIT5K} & \multicolumn{2}{c|}{SVT} & \multicolumn{3}{c|}{IC03}    & IC13-S & IC13-L  & SVT-P       & ICIST-L  \\
    \cline{2-13}
                                                     & 50   & 1K   & None          & 50        & None         & 50   & Full      & None      & None   & None  & None          & None   \\
    \hline
    ABBYY \cite{wang2011end}                         & 24.3 & -    & -             & 35.0      & -            & 56.0 & 55.0      & -         &   -    & -     & -             & -      \\
    \hline
    Wang \etal~\cite{wang2011end}                    & -    & -    & -             & 57.0      & -            & 76.0 & 62.0      & -         &   -    & -     & -             & -      \\
    \hline
    Mishra \etal~\cite{mishra2012scene}              & 64.1 & 57.5 & -             & 73.2      & -            & 81.8 & 67.8      & -         &   -    & -     & -             & -      \\
    \hline
    Wang \etal~\cite{wang2012end}                    & -    & -    & -             & 70.0      & -            & 90.0 & 84.0      & -         &   -    & -     & -             & -      \\
    \hline
    PhotoOCR \cite{bissacco2013photoocr}             & -    & -    & -             & 90.4      & 78.0         & -    & -         & -         &   -    & 87.6  & -             & -      \\
    \hline
    Phan \etal~\cite{phan2013recognizing}            & -    & -    & -             & 73.7      & -            & 82.2 & -         & -         &   -    & -     & -             & -      \\
    \hline
    Almaz{\'a}n \etal~\cite{almazan2014word}         & 91.2 & 82.1 & -             & 89.2      & -            & -    & -         & -         &   -    & -     & -             & -      \\
    \hline
    Lee \etal~\cite{lee2014region}                   & -    & -    & -             & 80.0      & -            & 88.0 & 76.0      & -         &   -    & -     & -             & -      \\
    \hline
    Yao \etal~\cite{yao2014strokelets}               & 80.2 & 69.3 & -             & 75.9      & -            & 88.5 & 80.3      & -         &   -    & -     & -             & -      \\
    \hline
    Jaderberg \etal~\cite{jaderberg14deepfeatures}   & -    & -    & -             & 86.1      & -            & 96.2 & 91.5      & -         &   -    & -     & -             & -      \\
    \hline
    Su \etal~\cite{su2014accurate}                   & -    & -    & -             & 83.0      & -            & 92.0 & 82.0      & -         &   -    & -     & -             & -      \\
    \hline
    Jaderberg \etal~\cite{Jaderberg15deepstructured} & 95.5 & 89.6 & -             & 93.2      & 71.7         & 97.8 & 97.0      & 89.6      &   -    & 81.8  & -             & -      \\
    \hline
    Jaderberg \etal~\cite{jaderberg2016reading}      & 97.1 & 92.7 & -             & 95.4      & 80.7*        & {\bf 98.7} & {\bf 98.6}      & {\bf 93.1*}     & -      & {\bf 90.8*} & -             & -      \\
    \hline
    $\text{R}^2\text{AM}$~\cite{lee2016recursive}    & 96.8 & 94.4 & 78.4          & 96.3      & 80.7         & 97.9 & 97.0      & 88.7      & 90.0   &  -         & -             & -      \\
    \hline
    CRNN \cite{shi2016end}                           & 97.8 & 95.0 & 81.2          &{\bf 97.5} & 82.7         & 98.7 &98.0       & 91.9      &   -    & 89.6       & 66.8          & -      \\
    \hline
    SRN \cite{shi2016robust}                         & 96.5 & 92.8 & 79.7          & 96.1      & 81.5         & 97.8 & 96.4      & 88.7      & 87.5   &  -         & -             & -      \\
    \hline
    RARE \cite{shi2016robust}                        & 96.2 & 93.8 & 81.9          & 95.5      & 81.9         & 98.3 & 96.2      & 90.1      & 88.6   &  -         & 71.8          & -      \\
    \hline
    STAR-Net \cite{liu16star}                        & 97.7 & 94.5 & 83.3          & 95.5      & 83.6         & 96.9 & 95.3      & 89.9      &   -    & 89.1       & 73.5          & -      \\
    \hline
    Char-Net \cite{liu18aaai}                        & -    & -    & 83.6          & -         & 84.4         &  -   &    -      & 91.5      &  90.8  & -          & 73.5          & 60     \\
    \hline
    \textbf{S-SAN}                                   &{\bf 98.4}&{\bf 96.1}&{\bf 85.2}& 97.1      & {\bf 85.5}   & 98.5 & 97.7      &92.9 &{\bf 91.1} & 90.3  & {\bf 74.4}&{\bf 65.7}\\
    \hline
\end{tabular}
\end{center}
\end{table*}

\subsection{Comparison with Other Methods}\label{subsection:comparative_evaluation}

The recognition accuracies of S-SAN are reported in Table~\ref{table:accuracies_on_six_benchmarks}. Compared with the recent deep-learning based methods \cite{shi2016robust,lee2016recursive,liu18aaai,shi2016end,liu16star,Jaderberg15deepstructured,jaderberg2016reading}, S-SAN can achieve state-of-the-art (or, in some cases, extremely comparative) performance on both constrained and unconstrained text recognition. 

In particular, we compare S-SAN against RARE \cite{shi2016robust}, STAR-Net \cite{liu16star} and Char-Net \cite{liu18aaai}, which are specifically designed for recognizing distorted text. In order to handle the distortion of text, RARE, STAR-Net and Char-Net all employ spatial transformer networks (STNs) \cite{jaderberg15spatial} in their feature encoders. From the recognition results in Table \ref{table:accuracies_on_six_benchmarks}, we find S-SAN outperforms all the three models by a large magain on almost every public benchmark. Even for the datasets IIIT5K and SVT-P that contain distorted text images, S-SAN without using any spatial transformer network can still achieve either extremely competitive or better performance. In order to explore the advantages of S-SAN comprehensively, we further report the complexity of these three models. In our implementation, S-SAN has 10.6 million parameters in total, while the encoders of RARE, STAR-Net and Char-Net already have about 16.5 million, 14.8 million and 12.8 million parameters respectively, as reported in Table~\ref{table:architecture}.

By comparing with RARE, STAR-Net and Char-Net, we can see that S-SAN is much simpler but can still achieve much better performance on recognizing distorted text. This is mainly because SAFE can effectively handle the problem of encoding characters with varying scales. On one hand, with the multi-scale convolutional encoder encoding the text images under multiple scales and the scale attention network automatically selecting features from the most relevant scale(s), SAFE can extract scale-invariant features from characters whose scales are affected by the distortion of the image. On the other hand, by explicitly tackling the scale problem, SAFE can put more attention on extracting discriminative features from distorted characters, which enables S-SAN a more robust performance when handling distortion of text images.

\begin{table*}[t]
\renewcommand{\arraystretch}{1.1}
\centering
\caption{Text recognition accuracies (\%) of models using more training data.}\label{table:accuracies_synthm}
\scriptsize
\begin{tabular}{|l|c|c|c|c|c|c|c|c|c|c|c|c|c|}
    \hline
    \multirow{2}*{Method}                    & \multicolumn{3}{c|}{IIIT5K} & \multicolumn{2}{c|}{SVT} & \multicolumn{3}{c|}{IC03}    & IC13-L      & SVT-P       & ICIST-S   & ICIST-L  \\
    \cline{2-13}
                                             & 50   & 1K   & None          & 50        & None         & 50   & Full      & None      & None      & None          & -         & None   \\
    \hline
    AON \cite{cheng2017arbitrarily}          &{\bf 99.6}& 98.1 & 87.0      & 96.0      & 82.8         & 98.5 & 97.1      & 91.5      & -         & 73.0          &  -        & 68.2   \\
    \hline
    FAN \cite{cheng2017focusing}             & 99.3 & 97.5 & 87.4          & 97.1      & 85.9         &{\bf 99.2}& 97.3  & 94.2      & 93.3      & -             &  70.6     & 66.2   \\
    \hline
    Bai \etal~\cite{bai2018edit}             & 99.5 & 97.9 & 88.3          & 96.6      & 87.5         & 98.7 & 97.9      & 94.6      &{\bf 94.4} & -             & 73.9      & -           \\
    \hline
    \textbf{S-SAN}                           & 99.0 & 97.9 & 90.5          & 97.1      & 87.0         & 98.5 & 97.6      & 93.7      & 91.9      & 77.1          & 76.9      & 70.7        \\
    \hline
    \textbf{S-SAN$_\text{res}$}              & 99.3 &{\bf 98.3}&\bf{91.5}  & {\bf 98.6}& {\bf 89.6}   & 99.1 &{\bf 98.0} &{\bf 94.9} & 93.8      & {\bf 81.6}    & \bf{80.0} & {\bf 73.4}   \\
    \hline
\end{tabular}
\end{table*}

Note that the previous state-of-the-art methods \cite{cheng2017focusing,bai2018edit} employ a 30-layer residual CNN with one BLSTM network in their encoders. We also train a deep version of our text recogniser (denoted as $\text{S-SAN}_\text{res}$), which employs a deep backbone CNN (denoted as $\text{SAFE}_\text{res}$ in Table 1). Besides, \cite{cheng2017focusing} and \cite{bai2018edit} were both trained using a 12-million dataset, which consists of 8-million images from SynthR and 4-million pixel-wise labeled images from \cite{gupta2016synthetic}. Following \cite{cheng2017focusing,bai2018edit}, we also train $\text{S-SAN}_\text{res}$ using the 12-million dataset. The results of S-SAN$_\text{res}$ are reported in Table \ref{table:accuracies_synthm}. As we can see from the table, S-SAN$_\text{res}$ achieves better results than \cite{cheng2017focusing,bai2018edit} on almost every benchmark. In particular, S-SAN$_\text{res}$ significantly outperforms \cite{cheng2017focusing,bai2018edit} on the most challenging dataset ICIST. Besides, unlike \cite{cheng2017focusing} which requires extra pixel-wise labeling of characters for training, S-SAN can be easily optimized using only the text images and their corresponding labels.

\section{Conclusion}
In this paper, we present a novel scale aware feature encoder (SAFE) to tackle the problem of having characters with different scales in the text images. SAFE is composed of a multi-scale convolutional encoder and a scale attention network. It can automatically extract scale-invariant features from characters with different scales. By explicitly handling the scale problem, SAFE can put more effort on handling other challenges in text recognition. Moreover, SAFE can transfer the learning of feature encoding across different character scales. This is particularly important for text recognizers to achieve a much more robust performance as it is nearly impossible to precisely control the scale distribution of characters in a training dataset. To demonstrate the effectiveness of SAFE, we design a simple but efficient scale-spatial attention network (S-SAN) for scene text recognition. Experiments on six public benchmarks illustrate that S-SAN can achieve state-of-the-art performance without any post-processing.

\bibliographystyle{splncs}
\bibliography{egbib}

\begin{thebibliography}{10}

\bibitem{bahdanau2014neural}
Bahdanau, D., Cho, K., Bengio, Y.:
\newblock Neural machine translation by jointly learning to align and
  translate.
\newblock arXiv preprint arXiv:1409.0473 (2014)

\bibitem{shi2016robust}
Shi, B., Wang, X., Lyu, P., Yao, C., Bai, X.:
\newblock Robust scene text recognition with automatic rectification.
\newblock In: IEEE Conference on Computer Vision and Pattern Recognition.
  (2016)

\bibitem{lee2016recursive}
Lee, C.Y., Osindero, S.:
\newblock Recursive recurrent nets with attention modeling for ocr in the wild.
\newblock In: IEEE Conference on Computer Vision and Pattern Recognition.
  (2016)

\bibitem{cheng2017focusing}
Cheng, Z., Bai, F., Xu, Y., Zheng, G., Pu, S., Zhou, S.:
\newblock Focusing attention: Towards accurate text recognition in natural
  images.
\newblock arXiv preprint arXiv:1709.02054v3 (2017)

\bibitem{liu18aaai}
Liu, W., Chen, C., Wong, K.Y.K.:
\newblock Char-net: A character-aware neural network for distorted scene text
  recognition.
\newblock In: AAAI Conference on Artificial Intelligence. (2018)

\bibitem{cheng2017arbitrarily}
Cheng, Z., Liu, X., Bai, F., Niu, Y., Pu, S., Zhou, S.:
\newblock Arbitrarily-oriented text recognition.
\newblock arXiv preprint arXiv:1711.04226 (2017)

\bibitem{shi2016end}
Shi, B., Bai, X., Yao, C.:
\newblock An end-to-end trainable neural network for image-based sequence
  recognition and its application to scene text recognition.
\newblock IEEE Transactions on Pattern Analysis and Machine Intelligence (2016)

\bibitem{liu16star}
Liu, W., Chen, C., Wong, K.K., Su, Z., Han, J.:
\newblock Star-net: {A} spatial attention residue network for scene text
  recognition.
\newblock In: British Machine Vision Conference. (2016)

\bibitem{luo2016understanding}
Luo, W., Li, Y., Urtasun, R., Zemel, R.:
\newblock Understanding the effective receptive field in deep convolutional
  neural networks.
\newblock In: Advances in Neural Information Processing Systems. (2016)

\bibitem{jaderberg14syntheticdata}
Jaderberg, M., Simonyan, K., Vedaldi, A., Zisserman, A.:
\newblock Synthetic data and artificial neural networks for natural scene text
  recognition.
\newblock In: Workshop on Deep Learning, Advances in Neural Information
  Processing Systems. (2014)

\bibitem{ijcai2017-458}
Yang, X., He, D., Zhou, Z., Kifer, D., Giles, C.L.:
\newblock Learning to read irregular text with attention mechanisms.
\newblock In: Twenty-Sixth International Joint Conference on Artificial
  Intelligence, {IJCAI-17}. (2017)

\bibitem{wang2010word}
Wang, K., Belongie, S.:
\newblock Word spotting in the wild.
\newblock In: European Conference on Computer Vision. (2010)

\bibitem{wang2011end}
Wang, K., Babenko, B., Belongie, S.:
\newblock End-to-end scene text recognition.
\newblock In: IEEE International Conference on Computer Vision. (2011)

\bibitem{neumann2012real}
Neumann, L., Matas, J.:
\newblock Real-time scene text localization and recognition.
\newblock In: IEEE Conference on Computer Vision and Pattern Recognition.
  (2012)

\bibitem{yao2014strokelets}
Yao, C., Bai, X., Shi, B., Liu, W.:
\newblock Strokelets: A learned multi-scale representation for scene text
  recognition.
\newblock In: IEEE Conference on Computer Vision and Pattern Recognition.
  (2014)

\bibitem{wang2012end}
Wang, T., Wu, D.J., Coates, A., Ng, A.Y.:
\newblock End-to-end text recognition with convolutional neural networks.
\newblock In: IEEE International Conference on Pattern Recognition. (2012)

\bibitem{bissacco2013photoocr}
Bissacco, A., Cummins, M., Netzer, Y., Neven, H.:
\newblock Photoocr: Reading text in uncontrolled conditions.
\newblock In: IEEE International Conference on Computer Vision. (2013)

\bibitem{jaderberg14deepfeatures}
Jaderberg, M., Vedaldi, A., Zisserman, A.:
\newblock Deep features for text spotting.
\newblock In: European Conference on Computer Vision.
\newblock (2014)

\bibitem{Jaderberg15deepstructured}
Jaderberg, M., Simonyan, K., Vedaldi, A., Zisserman, A.:
\newblock Deep structured output learning for unconstrained text recognition.
\newblock In: International Conference on Learning Representations. (2015)

\bibitem{jaderberg2016reading}
Jaderberg, M., Simonyan, K., Vedaldi, A., Zisserman, A.:
\newblock Reading text in the wild with convolutional neural networks.
\newblock International Journal of Computer Vision \textbf{116} (2016)  1--20

\bibitem{He2016ReadingST}
He, P., Huang, W., Qiao, Y., Loy, C.C., Tang, X.:
\newblock Reading scene text in deep convolutional sequences.
\newblock In: AAAI Conference on Artificial Intelligence. (2016)

\bibitem{graves2006connectionist}
Graves, A., Fern{\'a}ndez, S., Gomez, F., Schmidhuber, J.:
\newblock Connectionist temporal classification: labelling unsegmented sequence
  data with recurrent neural networks.
\newblock In: International Conference on Machine Learning. (2006)

\bibitem{jaderberg15spatial}
Jaderberg, M., Simonyan, K., Zisserman, A., Kavukcuoglu, K.:
\newblock Spatial transformer networks.
\newblock In: Advances in Neural Information Processing Systems. (2015)

\bibitem{chen2016attention}
Chen, L.C., Yang, Y., Wang, J., Xu, W., Yuille, A.L.:
\newblock Attention to scale: Scale-aware semantic image segmentation.
\newblock In: IEEE Conference on Computer Vision and Pattern Recognition.
  (2016)

\bibitem{bai2018edit}
Bai, F., Cheng, Z., Niu, Y., Pu, S., Zhou, S.:
\newblock Edit probability for scene text recognition.
\newblock arXiv preprint arXiv:1805.03384 (2018)

\bibitem{mishra2012scene}
Mishra, A., Alahari, K., Jawahar, C.:
\newblock Scene text recognition using higher order language priors.
\newblock In: British Machine Vision Conference. (2012)

\bibitem{lucas2005icdar}
Lucas, S.M., Panaretos, A., Sosa, L., Tang, A., Wong, S., Young, R., Ashida,
  K., Nagai, H., Okamoto, M., Yamamoto, H.,  et~al.:
\newblock Icdar 2003 robust reading competitions: entries, results, and future
  directions.
\newblock International Journal of Document Analysis and Recognition \textbf{7}
  (2005)  105--122

\bibitem{karatzas2013icdar}
Karatzas, D., Shafait, F., Uchida, S., Iwamura, M., Gomez~i Bigorda, L.,
  Robles~Mestre, S., Mas, J., Fernandez~Mota, D., Almazan~Almazan, J., de~las
  Heras, L.P.:
\newblock Icdar 2013 robust reading competition.
\newblock In: IEEE International Conference on Document Analysis and
  Recognition. (2013)

\bibitem{phan2013recognizing}
Phan, T., Shivakumara, P., Tian, S., Tan, C.:
\newblock Recognizing text with perspective distortion in natural scenes.
\newblock In: IEEE International Conference on Computer Vision. (2013)

\bibitem{karatzas2015icdar}
Karatzas, D., Gomez-Bigorda, L., Nicolaou, A., Ghosh, S., Bagdanov, A.,
  Iwamura, M., Matas, J., Neumann, L., Chandrasekhar, V.R., Lu, S.,  et~al.:
\newblock Icdar 2015 competition on robust reading.
\newblock In: IEEE International Conference on Document Analysis and
  Recognition. (2015)

\bibitem{ioffe2015batch}
Ioffe, S., Szegedy, C.:
\newblock Batch normalization: Accelerating deep network training by reducing
  internal covariate shift.
\newblock In: International Conference on Machine Learning. (2015)

\bibitem{zeiler2012adadelta}
Zeiler, M.D.:
\newblock Adadelta: an adaptive learning rate method.
\newblock arXiv preprint arXiv:1212.5701 (2012)

\bibitem{almazan2014word}
Almaz{\'a}n, J., Gordo, A., Forn{\'e}s, A., Valveny, E.:
\newblock Word spotting and recognition with embedded attributes.
\newblock IEEE Transactions on Pattern Analysis and Machine Intelligence
  \textbf{36} (2014)  2552--2566

\bibitem{lee2014region}
Lee, C.Y., Bhardwaj, A., Di, W., Jagadeesh, V., Piramuthu, R.:
\newblock Region-based discriminative feature pooling for scene text
  recognition.
\newblock In: IEEE Conference on Computer Vision and Pattern Recognition.
  (2014)

\bibitem{su2014accurate}
Su, B., Lu, S.:
\newblock Accurate scene text recognition based on recurrent neural network.
\newblock In: Asian Conference on Computer Vision.
\newblock (2014)

\bibitem{gupta2016synthetic}
Gupta, A., Vedaldi, A., Zisserman, A.:
\newblock Synthetic data for text localisation in natural images.
\newblock In: IEEE Conference on Computer Vision and Pattern Recognition.
  (2016)

\end{thebibliography}

\end{document}